\pdfoutput=1
\documentclass[10pt,a4paper]{article}

\usepackage[utf8]{inputenc}
\usepackage[T1]{fontenc}
\usepackage{lmodern}
\usepackage[margin=0.9in]{geometry}
\usepackage{graphicx}
\usepackage{booktabs}
\usepackage{array}
\usepackage{hyperref}
\usepackage{xcolor}
\usepackage{amsmath}
\usepackage{float}
\usepackage{caption}
\usepackage{enumitem}
\usepackage{parskip}
\usepackage{titlesec}
\usepackage{fancyhdr}
\usepackage{authblk}
\usepackage{footmisc}

\hypersetup{colorlinks=true, linkcolor=blue!60!black, citecolor=blue!60!black, urlcolor=blue!60!black}
\titleformat{\section}{\normalsize\bfseries}{\thesection}{0.5em}{}
\titleformat{\subsection}{\small\bfseries}{\thesubsection}{0.4em}{}
\titlespacing*{\section}{0pt}{10pt}{3pt}
\titlespacing*{\subsection}{0pt}{6pt}{2pt}
\renewenvironment{abstract}{\section*{Abstract}}{\par\medskip}

\title{%
    \vspace{-1.2cm}
    \textbf{Do Thought Streams Matter?\\[2pt]
    Evaluating Reasoning in Gemini Vision-Language Models for Video Scene Understanding}%
    \vspace{-0.4cm}
}

\author{Shivam Sharma}
\author{Sankalp Nagaonkar}
\author{Ashish Choithani}
\author{Ashutosh Trivedi}

\affil{
VideoDB\thanks{
\url{https://videodb.io}\par
Code: \url{https://github.com/video-db/gemini-reasoning-eval}
}
}
\affil{\texttt{engg@videodb.io}}

\date{}

\begin{document}

\maketitle
\thispagestyle{fancy}
% ==========================================================================
% ABSTRACT
% ==========================================================================
\begin{abstract}
\noindent We benchmark how internal reasoning traces that we call \emph{thought streams} affects video scene understanding in vision-language models. Using four configurations of Google's Gemini 2.5 Flash and Flash Lite across scenes extracted from 100 hours of video, we ask three questions: Does more thinking lead to better outputs? Where do the gains stop? And what do these models actually think about?

We introduce three evaluation metrics: \emph{Contentfulness} measures how much of the thought stream is useful scene content versus meta-commentary; \emph{Thought--Final Coverage} measures how faithfully the thought stream translates into the final output; and \emph{Dominant Entity Analysis} identifies which subjects, actions, and settings the model focuses on. GPT-5 serves as an independent judge.

Key findings: quality gains from additional thinking plateau quickly, with most improvement happening in the first few hundred tokens. Flash Lite offers the best balance between quality and token usage. Tight reasoning budgets cause the model to add content in the final output that it never reasoned about, a form of compression-step hallucination. Despite being different model tiers, Flash and Flash Lite produce similar thought streams, though they differ in style: Flash discusses its reasoning process, while Lite focuses on describing the actual scene.

% \noindent Code: \url{https://github.com/video-db/gemini-reasoning-eval}
\end{abstract}

% ==========================================================================
\section{Introduction}
% ==========================================================================

Vision--language models (VLMs) are increasingly used for structured video understanding, extracting subjects, actions, settings, and emotions from video scenes at scale. Recent models like Gemini 2.5 support \emph{extended thinking}: the model generates an internal chain of thought (a ``thought stream'') before producing its final structured output.

This raises practical questions. Does the thought stream actually contain useful information, or is it mostly filler? Does the final output faithfully reflect what was reasoned about? And do different thinking budgets change \emph{what} the model pays attention to? For example, does a model with a small budget default to generic labels like ``person'' instead of identifying specific subjects?

At VideoDB, we process large volumes of video content and need clear answers to these questions for production deployment. Existing benchmarks evaluate only the final output and treat the model as a black box. We take a different approach: we look inside the thought stream, compare it to the final output, and measure what gets lost or added along the way.

\textbf{Contributions.} (1)~A benchmark framework with three metrics that measure thought stream quality, compression fidelity, and attentional focus. (2)~Evaluation of four Gemini 2.5 variants across scenes from 100 hours of video, with token cost analysis from the Gemini API. (3)~Evidence that Flash and Flash Lite produce nearly identical thought streams despite being different model tiers.

% ==========================================================================
\section{Related Work}
% ==========================================================================

\textbf{Model benchmarking.} MMLU~\cite{hendrycks2021mmlu} evaluates broad knowledge across 57 academic subjects using multiple-choice questions, testing how well models generalize across domains. HellaSwag~\cite{zellers2019hellaswag} tests commonsense reasoning by asking models to choose the most plausible continuation of everyday scenarios, using adversarial filtering to create challenging distractors. HumanEval~\cite{chen2021humaneval} measures functional code generation by asking models to complete Python functions and checking them against unit tests. Industry reports from Google~\cite{gemini2023}, OpenAI~\cite{openai2023gpt4}, and Meta~\cite{touvron2023llama} extend these evaluations to multi-modal tasks including image and video understanding. We complement these benchmarks by looking inside the reasoning process, not just the final answer.

\textbf{Chain-of-thought.} Wei et al.~\cite{wei2022cot} showed that adding intermediate reasoning steps (``let's think step by step'') to prompts significantly improves performance on arithmetic, commonsense, and symbolic reasoning tasks, particularly for large models. Their work demonstrated that reasoning ability emerges at scale and can be elicited through prompting alone. We go further by measuring the \emph{quality} of the chain of thought: how much of it is useful content, and how much survives into the final output.

\textbf{Video understanding.} ActivityNet~\cite{caba2015activitynet} provides 20,000 untrimmed YouTube videos annotated with 200 activity classes, enabling temporal action detection and dense captioning at scale. Ego4D~\cite{grauman2022ego4d} offers 3,670 hours of first-person video across 74 locations worldwide, targeting tasks like episodic memory, forecasting, and hand--object interaction from an egocentric perspective. Video-MME~\cite{fu2024videomme} is the first comprehensive benchmark for evaluating multi-modal LLMs on video analysis, covering short to long videos with both multiple-choice and open-ended questions across diverse domains. Our benchmark focuses on structured metadata extraction (subjects, actions, settings, emotions), which is closer to real production needs.

\textbf{LLM-as-judge.} Zheng et al.~\cite{zheng2023llmjudge} introduced MT-Bench and Chatbot Arena, demonstrating that strong LLMs (such as GPT-4) can serve as scalable and reliable evaluators that closely approximate human preference rankings. Their work showed that LLM judges achieve over 80\% agreement with human annotators, making them practical alternatives to expensive human evaluation. We use GPT-5 as an extraction judge for computing coverage metrics.

% ==========================================================================
\section{Methodology}
% ==========================================================================

\subsection{Task and Dataset}

We use VideoDB to segment videos into individual scenes based on visual and semantic boundaries. Each scene is then processed independently through a VLM. Given video frames from a single scene, the VLM produces: (1)~an internal thought stream (the model's reasoning), and (2)~a structured JSON output containing extracted metadata including subjects, actions, settings, emotions, shot types, and more.

We treat the thought stream as an observable trace, not a complete record of internal computation. Our analysis focuses on how this trace relates to the final output.

Our dataset comprises approximately 100 hours of video spanning 37 distinct visual styles, including 2D and 3D animation, cinematic and documentary footage, gameplay and esports recordings, live concert and event captures, motion graphics, surveillance, social media clips, vlogs, and vintage film. Content categories cover 38 domains such as entertainment, sports, news, educational, culinary, music performance, drama, comedy, reality television, gaming, corporate, travel vlog, and children's content, among others. In terms of production quality, 63.63\% of the videos are classified as high quality, 34.09\% as medium, and 2.28\% as low, reflecting a distribution weighted toward professionally produced content.

Frames are extracted at 1~FPS with a cap of 10 frames per scene. Each scene is processed independently with no cross-scene context. Across all four model variants, this produces over 93,000 scene-level results in total.

\subsection{Models Under Evaluation}

We evaluate two Gemini 2.5 model tiers, Flash and Flash Lite, across four reasoning budget configurations (Table~\ref{tab:variants}). All variants receive identical prompts, schemas, and temperature settings. The only difference is the thinking budget: how many tokens the model is allowed to spend on its thought stream before producing the final output.

Table~\ref{tab:variants} shows the token breakdown per variant, computed from the Gemini API response fields across all scenes. Input tokens (the prompt text plus image frames) vary per scene depending on the number of frames extracted (1--10 at 1~FPS), but remain consistent across variants for the same scene since every variant sees the same frames. The key variable is thought tokens.

\begin{table}[H]
\centering
\small
\caption{Model variants and mean token usage per scene. Thought tokens are from the Gemini API; Input, Output, and Total are estimates.}
\label{tab:variants}
\begin{tabular}{@{}llrrrrr@{}}
\toprule
\textbf{Variant} & \textbf{Tier} & \textbf{Budget} & \textbf{Thought} & \textbf{Input} & \textbf{Output} & \textbf{Total} \\
\midrule
Flash -- 128         & Flash      & 128 tokens   & 105   & 1,964 & 262 & 2,331 \\
Flash -- Dynamic     & Flash      & Unlimited    & 1,021 & 1,978 & 259 & 3,258 \\
\midrule
Lite -- 512          & Flash Lite & 512 tokens   & 366   & 1,976 & 222 & 2,563 \\
Lite -- 1024         & Flash Lite & 1,024 tokens & 718   & 1,976 & 225 & 2,918 \\
\bottomrule
\end{tabular}
\end{table}

% Input tokens vary per scene depending on the number of frames (1--10 at 1~FPS), with a fixed text prompt component and a variable image token component that scales with frame count. Output tokens (the structured JSON response, excluding thought tokens) are relatively stable across variants. The main cost driver is the thought token budget.

\subsection{Evaluation Metrics}

We evaluate the thought stream and its relationship to the final output using three metrics. GPT-5 is used as a judge for the coverage metrics; contentfulness is fully deterministic.

\textbf{Contentfulness} $\in [0,1]$: measures what fraction of the thought stream consists of actual scene-related content (nouns and verbs describing the scene) versus meta-commentary (phrases like ``let me analyze'' or ``I need to think about''). Sentences containing meta-commentary are identified and removed using regex patterns (matching phrases like ``I will,'' ``let me,'' ``step by step,'' ``json,'' etc.). The remaining sentences are then POS-tagged using NLTK, and only words belonging to noun phrases or verb phrases are counted as content words.

\emph{Example.} Consider a thought stream: ``Let me analyze this scene carefully. A young woman sits at a wooden desk, typing on a silver laptop in a bright office.'' The first sentence (``Let me analyze this scene carefully'') is meta-commentary and gets filtered out. From the remaining text, content words are: \emph{woman, sits, desk, typing, laptop, office}. If the full trace has 20 total words and 6 are content words, the contentfulness score is $6/20 = 0.30$.

\textbf{Thought--Final Coverage (Thought Coverage and Output Grounding)} $\in [0,1]$: measures how well the thought stream and the final output align with each other. GPT-5 extracts atomic items (individual facts) from both the thought stream and the final JSON output. These items are matched using cascaded fuzzy matching (exact match $\rightarrow$ token-sort ratio $\geq 75$ $\rightarrow$ partial ratio $\geq 75$). The \emph{token-sort ratio} tokenizes both strings, and then computes the similarity of the sorted sequences. This handles cases where the same words appear in a different order (e.g., ``wooden desk'' vs.\ ``desk, wooden''). The \emph{partial ratio} finds the best matching substring of the shorter string within the longer one, handling cases where one item is a substring or slight rewording of another (e.g., ``laptop'' matching ``silver laptop'').

\emph{Thought Coverage} answers: of everything the model thought about, how much made it into the final output? Low thought coverage means the model reasoned about details but then dropped them.

\emph{Output Grounding} answers: of everything in the final output, how much was actually present in the thought stream? Low output grounding means the model added content not explicitly present in the thought stream. We call this \emph{compression-step hallucination}, a mismatch between the verbalized trace and the final output.

\emph{Example.} Suppose the thought stream mentions: [woman, desk, laptop, office, typing, focused expression]. The final output contains: [woman, desk, laptop, office, typing, smiling]. Five of six thought items appear in the output, so Thought Coverage = 5/6. Five of six output items were in the thought stream (``smiling'' was not), so Output Grounding = 5/6. $F1 = 2 \times (5/6 \times 5/6) / (5/6 + 5/6) \approx 0.83$.

\textbf{F1 Score}: $F1 = 2 \cdot TC \cdot OG\,/\,(TC + OG)$, where $TC$ = Thought Coverage and $OG$ = Output Grounding. Our metrics measure internal consistency between the thought stream and the final output, not correctness against external ground truth. A model can achieve high alignment while still being factually incorrect. We use these metrics to study how reasoning traces translate into outputs, rather than to evaluate absolute task accuracy.

\textbf{Dominant Entity Analysis}: for each scene, we identify the single most prominent subject, action, and setting. This lets us compare what different variants pay attention to. For example, does a low-budget model default to calling everyone ``person,'' or does it identify specific subjects like ``chef'' or ``streamer''? We track how these dominant entities shift as the thinking budget changes.

% ==========================================================================
\section{Results}
% ==========================================================================

\subsection{Main Quality Results}

Table~\ref{tab:main} presents the core evaluation metrics across all variants. Lite 1024 leads with the highest F1, Output Grounding, and Thought Coverage. Flash Dynamic is a close second. Flash 128 shows a large Output Grounding gap: roughly one in four of its output items were never mentioned in its thought stream.

\begin{table}[H]
\centering
\small
\caption{Main evaluation results. T.Cov.\ = Thought Coverage, O.Grd.\ = Output Grounding. Bold = best among thinking variants. Err.\% = fraction of scenes with processing errors.}
\label{tab:main}
\begin{tabular}{@{}lrrrrrr@{}}
\toprule
\textbf{Variant} & \textbf{Scenes} & \textbf{Cont.} & \textbf{T.Cov.} & \textbf{O.Grd.} & \textbf{F1} & \textbf{Err.\%} \\
\midrule
Flash -- 128        & 25,807 & .323 & .853 & .767 & .830 & 0.14 \\
Flash -- Dynamic    & 22,519 & \textbf{.594} & .953 & .964 & .957 & 0.13 \\
\midrule
Lite -- 512         & 23,091 & .520 & .940 & .948 & .942 & \textbf{0.09} \\
\textbf{Lite -- 1024} & 22,188 & .582 & \textbf{.954} & \textbf{.966} & \textbf{.959} & 0.12 \\
\bottomrule
\end{tabular}
\end{table}

\subsection{Token Cost Breakdown}

\begin{figure}[H]
\centering
\includegraphics[width=\textwidth]{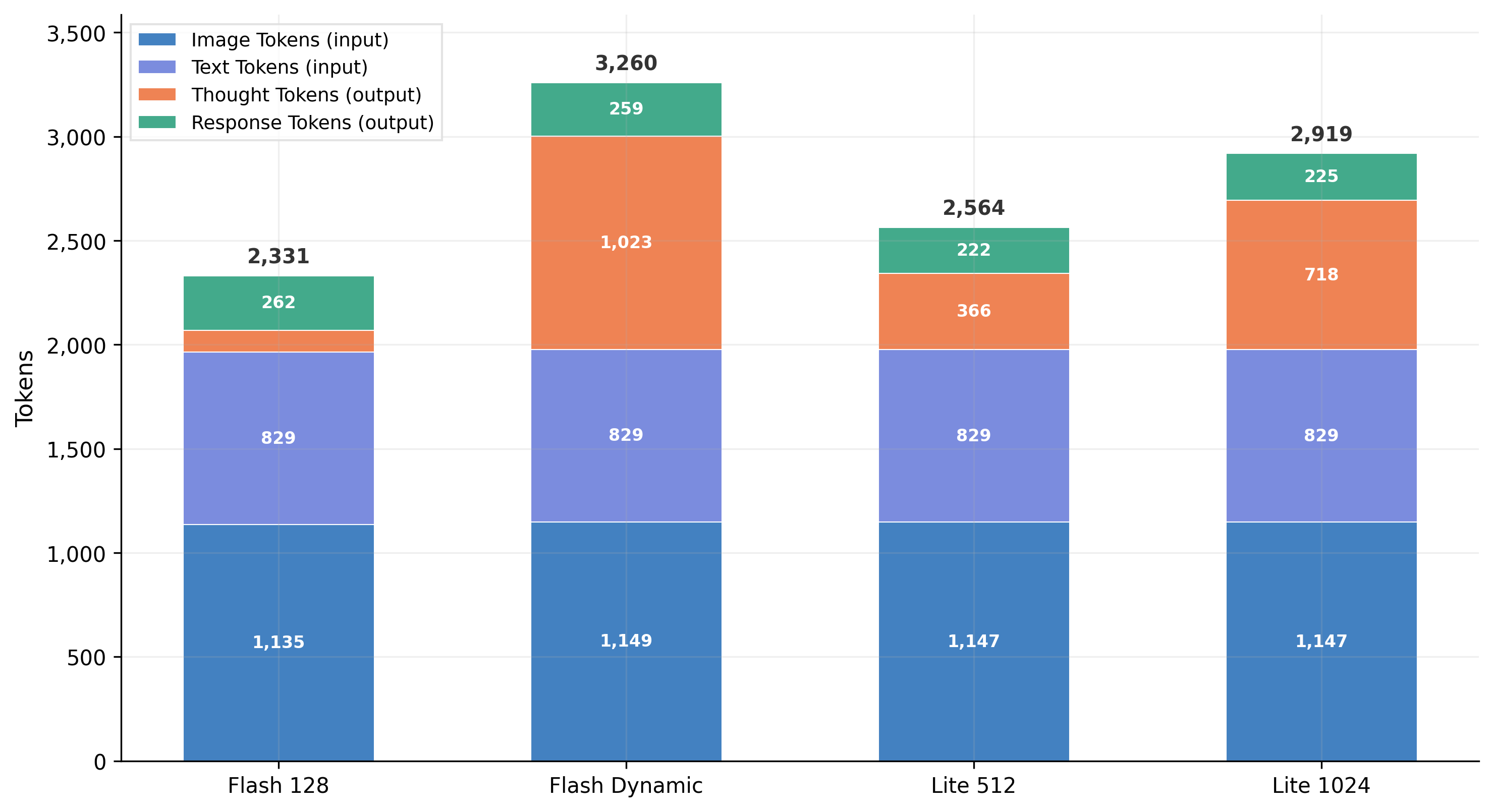}
\caption{Mean token breakdown per scene from the Gemini API. Input tokens (blue + purple) vary with frame count. Thought tokens (orange) are the main cost variable. Response tokens (green) are relatively stable.}
\label{fig:tokens}
\end{figure}

Figure~\ref{fig:tokens} shows how tokens are distributed across the four thinking variants. Input tokens (text prompt + image frames) vary per scene depending on frame count, but are consistent across variants for the same scene. The variable cost comes primarily from thought tokens.

Lite variants are cheaper than Flash variants at comparable quality levels. Lite 1024 uses an estimated mean of ${\sim}$2,918 total tokens per scene versus ${\sim}$3,258 for Flash Dynamic, while achieving equal or better F1. The saving comes from two places: Lite uses fewer thought tokens (718 vs.\ 1,021) and produces slightly shorter responses.

Flash 128 is the cheapest thinking variant at ${\sim}$2,331 tokens per scene, but this comes at a steep quality cost (F1 = 0.830 vs.\ 0.959 for Lite 1024). Lite 512 offers a much better trade-off: ${\sim}$2,563 tokens per scene with F1 = 0.942.

\subsection{Scaling and Diminishing Returns}

\begin{figure}[!htbp]
\centering
\includegraphics[width=\textwidth]{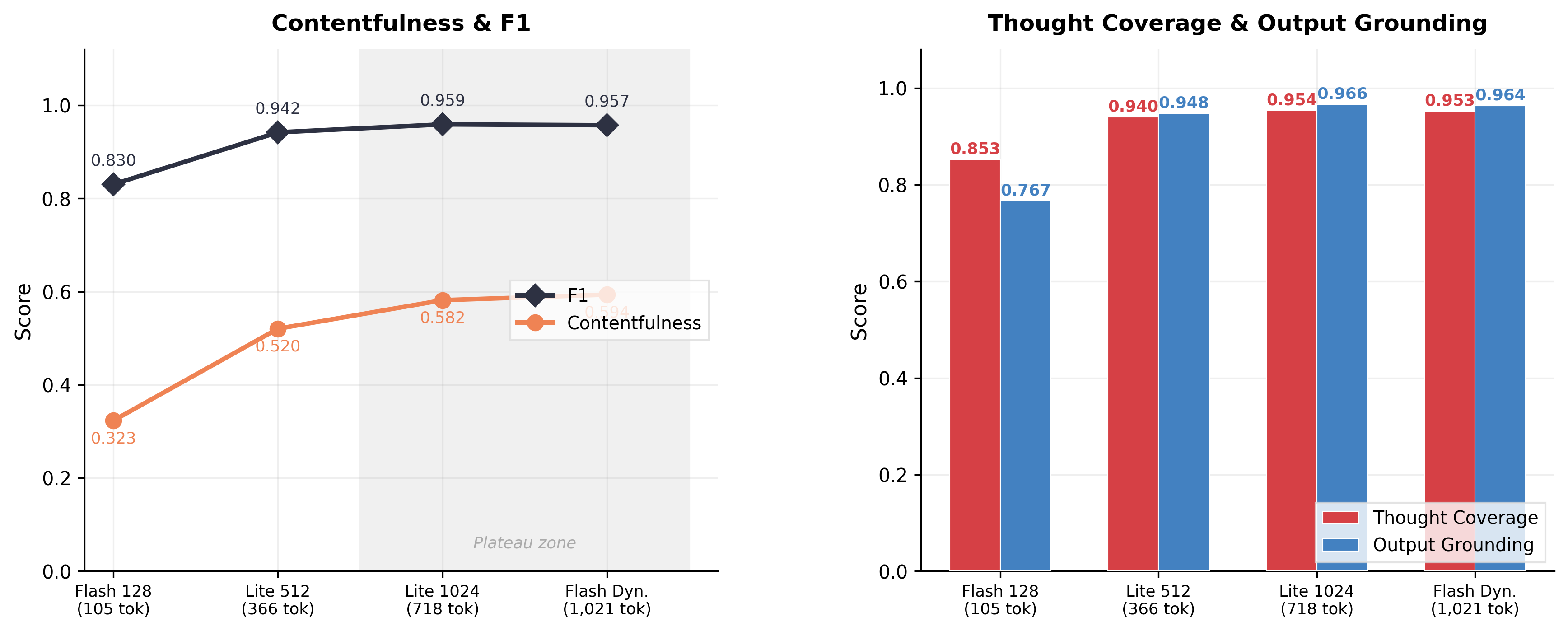}
\caption{Metric scaling with reasoning token budget. Left: Contentfulness rises steadily while F1 shows steep gains early then plateaus. Right: Thought Coverage and Output Grounding per variant; the large gap at Flash 128 reflects compression-step hallucination.}
\label{fig:scaling}
\end{figure}

Within the evaluated configurations, quality gains from additional thinking tokens show diminishing returns. Figure~\ref{fig:scaling} shows that contentfulness scales roughly linearly with the thinking budget, but F1 rises steeply in the first few hundred tokens and then flattens.

The jump from Flash 128 (105 mean thought tokens) to Lite 512 (366 tokens) yields a large F1 improvement. But going from Lite 512 to Lite 1024 (718 tokens), which nearly doubles the budget, gains only a small additional improvement. And Flash Dynamic (1,021 tokens) performs slightly below Lite 1024 despite using far more thought tokens. This pattern is consistent in our dataset and model settings; it may differ for other tasks or longer-horizon video reasoning.

\subsection{Consistency and Quality Tiers}

Averages can hide important variation. Table~\ref{tab:consistency} breaks down the quality distribution: what fraction of scenes achieve a perfect score, and what fraction fall below an acceptable threshold.

\begin{table}[H]
\centering
\small
\caption{Quality distribution. CV = coefficient of variation, computed as the ratio of standard deviation to the mean ($\text{CV} = \sigma / \mu$); lower values indicate more consistent performance across scenes. Perfect = F1 of 1.0. Low = F1 below 0.5.}
\label{tab:consistency}
\begin{tabular}{@{}lrrrrr@{}}
\toprule
\textbf{Variant} & \textbf{F1 Mean} & \textbf{F1 Std} & \textbf{CV} & \textbf{Perfect\%} & \textbf{Low\%} \\
\midrule
Flash -- 128      & 0.830 & 0.234 & 0.282 & 36.5 & 11.0 \\
Flash -- Dynamic  & 0.957 & 0.079 & 0.082 & 62.0 &  0.3 \\
Lite -- 512       & 0.942 & 0.097 & 0.103 & 55.8 &  0.6 \\
\textbf{Lite -- 1024} & \textbf{0.959} & \textbf{0.078} & \textbf{0.082} & \textbf{64.3} & \textbf{0.2} \\
\bottomrule
\end{tabular}
\end{table}

% Lite 1024 achieves the highest rate of perfect scores: nearly two-thirds of all scenes get a flawless alignment between thought stream and output. Flash 128 has over 50 times more failure scenes than Lite 1024 (11.0\% vs.\ 0.2\% with F1 below 0.5). For production systems where tail-end failures require human review, this gap is critical.

% \begin{figure}[H]
% \centering
% \includegraphics[width=\textwidth]{./figures/fig03_quality_tiers.png}
% \caption{Quality tier breakdown across variants. Left: perfect-score rate. Right: failure rate (F1 below 0.5). The gap between Flash 128 and the other variants is large.}
% \label{fig:tiers}
% \end{figure}

\subsection{Thought Stream Similarity}

An unexpected finding is that Flash and Flash Lite produce remarkably similar thought streams despite being different model tiers. Pairwise similarity scores between thought streams, computed by having an LLM judge compare the content of corresponding thought traces scene by scene, average around 0.88--0.90.

\begin{table}[H]
\centering
\small
\caption{Thought stream similarity scores (0--1) between model variants, measured by LLM-as-judge pairwise comparison.}
\label{tab:similarity}
\begin{tabular}{@{}llr@{}}
\toprule
\textbf{Variant A} & \textbf{Variant B} & \textbf{Similarity} \\
\midrule
\multicolumn{3}{@{}l}{\emph{Cross-tier (Flash $\leftrightarrow$ Lite)}} \\
Lite 1024 & Flash Dynamic & 0.887 \\
Lite 512  & Flash Dynamic & 0.881 \\
Lite 1024 & Flash 128     & 0.885 \\
Lite 512  & Flash 128     & 0.880 \\
\midrule
\multicolumn{3}{@{}l}{\emph{Within-tier}} \\
Lite 1024 & Lite 512       & 0.905 \\
Flash 128 & Flash Dynamic  & 0.903 \\
\midrule
\multicolumn{3}{@{}l}{\emph{Same model, repeated runs (determinism test)}} \\
Flash Dynamic & Flash Dynamic (rerun) & 0.893 \\
\bottomrule
\end{tabular}
\end{table}

There are two important takeaways. First, cross-tier similarity ($\sim$0.88) is nearly as high as within-tier similarity ($\sim$0.90), meaning Flash and Lite think about the same things. The qualitative difference is in \emph{style}: Flash tends to narrate its reasoning process (``Let me think about what's in this image...''), while Lite jumps straight to describing the scene content. This explains why Lite achieves comparable contentfulness with fewer tokens, spending a higher fraction of its budget on scene content rather than process narration.

Second, the determinism test, in which Flash Dynamic was run twice on the same videos, shows a similarity of 0.893. The areas that remain stable across runs include brand names, logos, text detection, and object identity. The areas that vary include emotion labels and location descriptions.

\subsection{Flash vs.\ Flash Lite}

\begin{figure}[H]
\centering
\includegraphics[width=\textwidth]{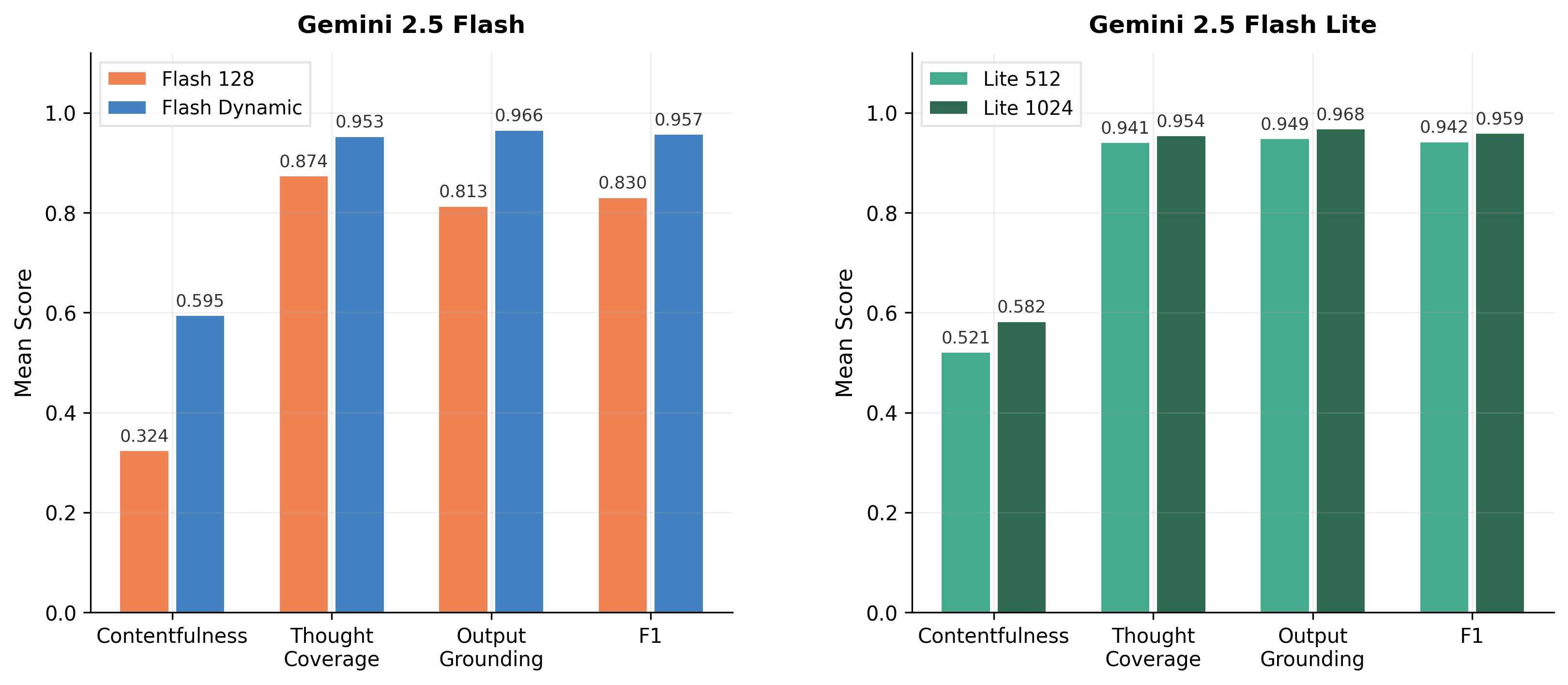}
\caption{Flash (left) vs.\ Flash Lite (right). Higher budgets improve all metrics. Lite matches or exceeds Flash quality at lower cost.}
\label{fig:family}
\end{figure}

Figure~\ref{fig:family} compares the two model tiers side by side. Within each tier, higher budgets improve all metrics.

Flash Lite 1024 matches or exceeds Flash Dynamic on every metric while using 30\% fewer thought tokens (718 vs.\ 1,021). In our setup, Lite shows better token efficiency. One possible reason is that it spends more of its budget on scene content and less on process narration.

% ==========================================================================
\section{Analysis}
% ==========================================================================

\subsection{Compression-Step Hallucination}

Flash 128's low Output Grounding (0.767) indicates that roughly one in four output items were never mentioned in its thought stream. We use \emph{compression-step hallucination} for cases where the final output includes information not explicitly present in the generated thought stream. This does not mean the model did not consider it internally; it signals a mismatch between the verbalized trace and the structured output.

Increasing the thinking budget significantly reduces this issue and leads to higher output grounding. With a more generous token budget, the model builds a sufficiently detailed reasoning trace that the output generation step can draw from, reducing the need to introduce content that was not previously reasoned about.

\subsection{Subject Specificity}

The Dominant Entity Analysis metric tracks the most prominent subject, action, and setting per scene. One clear signal is subject specificity: lower thinking budgets tend to produce more generic labels. Flash 128 assigns ``person'' as the dominant subject in about 15\% of scenes, compared to roughly 8\% for Flash Dynamic. A similar pattern appears within the Lite tier: 13\% at 512 tokens versus 11\% at 1024. As the thinking budget increases, models are more likely to identify specific subjects (``streamer,'' ``chef,'' ``cat'') rather than defaulting to safe, generic terms.

% ==========================================================================
\section{Limitations and Future Work}
% ==========================================================================

Overall, thought streams are best treated as observable traces, and alignment is best treated as an internal consistency signal rather than a correctness signal. In this work, we use these signals as a lens to study how models allocate reasoning effort and how that effort translates into structured outputs. The findings are intended as relative comparisons within this framework.

\textbf{Limitations.} Our metrics capture thought-to-output consistency, not correctness against human ground truth, so high alignment can still be wrong. Coverage is measured with a single LLM judge, which may introduce systematic bias. The dataset is limited to scene-level analysis at 1~FPS with up to 10 frames per scene, so it does not test long-range temporal reasoning, multi-scene dependencies, or narrative structure.

\textbf{Future work.} A natural next step is to pair these internal consistency metrics with human-annotated ground truth, separating correctness from self-consistency and measuring when better reasoning traces translate into real task-accuracy gains. We also plan extension to other VLMs (OpenAI, Anthropic, Open-source models); a fine-grained budget sweep from 64 to 2048 tokens; latency and dollar-cost profiling; domain-specific evaluation (medical, sports, surveillance); and larger-scale determinism testing.
% ==========================================================================
\section{Conclusion}
% ==========================================================================

We benchmarked four Gemini 2.5 configurations on video scene understanding, looking inside the thought stream rather than treating the model as a black box. The main takeaways:

\begin{enumerate}[nosep,leftmargin=*]
    \item \textbf{More thinking helps, but gains plateau quickly in our setup.} Most quality improvement happens in the first few hundred thought tokens. Beyond about 700 tokens, additional thinking adds cost with smaller gains in this dataset.
    \item \textbf{Lite 1024 is the quality leader.} It achieves the best F1, Thought Coverage, Output Grounding, and perfect-score rate while using 30\% fewer thought tokens than Flash Dynamic.
    % \item \textbf{Lite 512 is the best default for production.} It delivers strong quality at the lowest error rate and roughly half the thought cost of Lite 1024.
    \item \textbf{Tight budgets increase compression-step hallucination.} Flash 128 more often outputs details that were not explicitly present in its thought stream.
    \item \textbf{Flash and Lite think about the same things.} Cross-tier thought stream similarity is nearly as high as same-model determinism, suggesting the two tiers share underlying reasoning patterns.
    \item \textbf{Flash Lite is more token-efficient in this setup.} It tends to spend less on process narration and more on scene content.
\end{enumerate}

\bibliographystyle{plain}
\bibliography{references}

\end{document}